\documentclass[conference]{IEEEtran}
\IEEEoverridecommandlockouts
\usepackage{cite}
\usepackage{amsmath,amssymb,amsfonts}
\usepackage{algorithmic}
\usepackage{graphicx}
\usepackage{textcomp}
\usepackage{xcolor}
\usepackage{subcaption}
\usepackage{adjustbox}
\usepackage{wrapfig} 
\usepackage{booktabs} 
\usepackage{multirow} 
\usepackage{bm}

\def\BibTeX{{\rm B\kern-.05em{\sc i\kern-.025em b}\kern-.08em
    T\kern-.1667em\lower.7ex\hbox{E}\kern-.125emX}}
\begin{document}

\title{Enhanced Fracture Diagnosis Based on Critical Regional and Scale Aware in YOLO
}

\author{\IEEEauthorblockN{1\textsuperscript{st} Yuyang Sun}
\IEEEauthorblockA{\textit{Huazhong Agricultural University} \\
Wuhan, China \\
frontsea040320@gmail.com}
\and
\IEEEauthorblockN{2\textsuperscript{nd} Junchuan Yu}
\IEEEauthorblockA{\textit{Tianjin University} \\
Tianjin, China \\
hogwartsniffler@163.com}
\and
\IEEEauthorblockN{* Cuiming Zou\thanks{* Corresponding Author}}
\IEEEauthorblockA{\textit{Huazhong Agricultural University} \\
Wuhan, China \\
zoucuiming2006@163.com}
}
\maketitle

\begin{abstract}
Fracture detection plays a critical role in medical imaging analysis, traditional fracture diagnosis relies on visual assessment by experienced physicians, however the speed and accuracy of this approach are constrained by the expertise. With the rapid advancements in artificial intelligence, deep learning models based on the YOLO framework have been widely employed for fracture detection, demonstrating significant potential in improving diagnostic efficiency and accuracy. This study proposes an improved YOLO-based model, termed Fracture-YOLO, which integrates novel Critical-Region-Selector Attention (CRSelector) and Scale-Aware (ScA) heads to further enhance detection performance.
Specifically, the CRSelector module utilizes global texture information to focus on critical features of fracture regions. Meanwhile, the ScA module dynamically adjusts the weights of features at different scales, enhancing the model's capacity to identify fracture targets at multiple scales. Experimental results demonstrate that, compared to the baseline model, Fracture-YOLO achieves a significant improvement in detection precision, with mAP$_{50}$ and mAP$_{50-95}$ increasing by 4 and 3, surpassing the baseline model and achieving state-of-the-art (SOTA) performance.
\end{abstract}

\section{Introduction}
 A mid the global strain on healthcare resources, smart healthcare has emerged as an innovative development model, garnering widespread attention. Smart healthcare integrates advanced sensor and artificial intelligence (AI) technologies \cite{shaik2024survey,mridha2023interpretable,kothamali2023smart} to enable intelligent medical diagnostics and optimized management. X-ray-based fracture detection is particularly critical, as timely and accurate diagnosis of fractures is essential for patient outcomes, especially in regions with limited healthcare resources or a shortage of specialized physicians.

As AI continues to advance, computer vision technology is being increasingly applied in the field of medical imaging. Automatic identification and monitoring of fracture regions in X-rays using object detection technology has become an important research direction and has been used in a large number of practical applications in the clinic.
\begin{figure}[htbp]  
	\centering
	\begin{minipage}{0.115\textwidth}  
		\centering
		\includegraphics[width=\columnwidth]{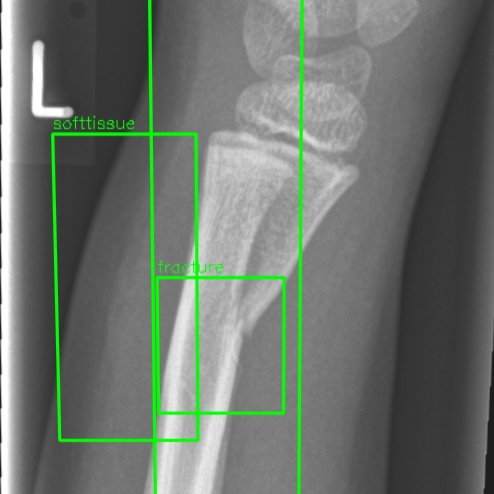}
		\caption*{(a)}
		\label{fig:1a}
	\end{minipage}%
	\hfill
	\begin{minipage}{0.115\textwidth}
		\centering
		\includegraphics[width=\columnwidth]{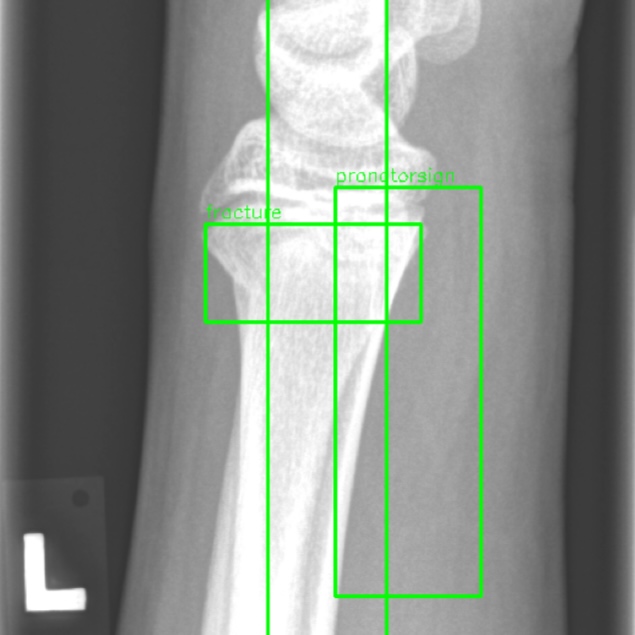}
		\caption*{(b)}
		\label{fig:1b}
	\end{minipage}%
	\hfill
	\begin{minipage}{0.115\textwidth}
		\centering
		\includegraphics[width=\columnwidth]{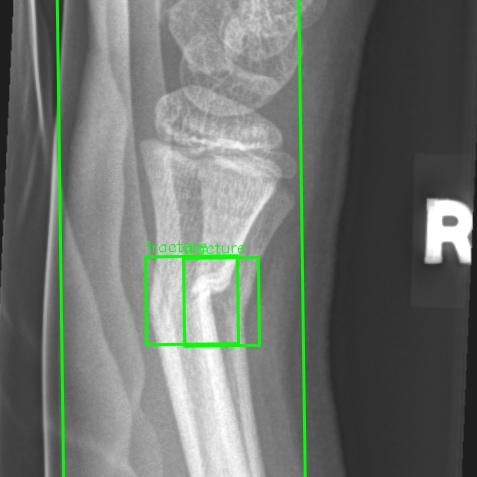}
		\caption*{(c)}
		\label{fig:1c}
	\end{minipage}%
	\hfill
	\begin{minipage}{0.115\textwidth}
		\centering
		\includegraphics[width=\columnwidth]{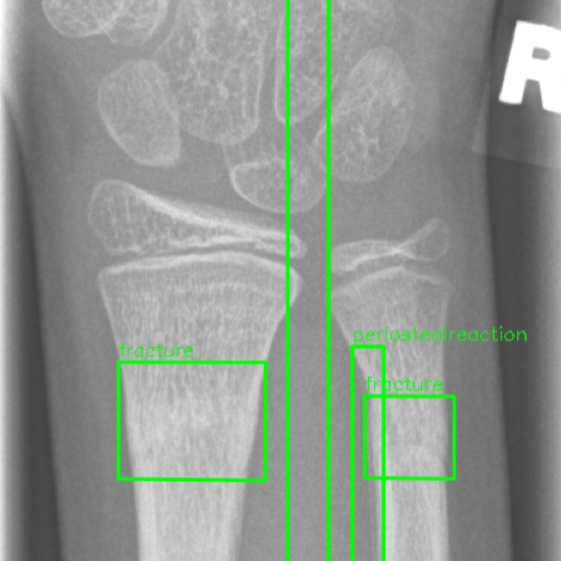}
		\caption*{(d)}
		\label{fig:1d}
	\end{minipage}
	\caption{Sample labeled X-ray images.}
\end{figure}
\begin{figure*}[h]
	\centering
	\includegraphics[width=\textwidth]{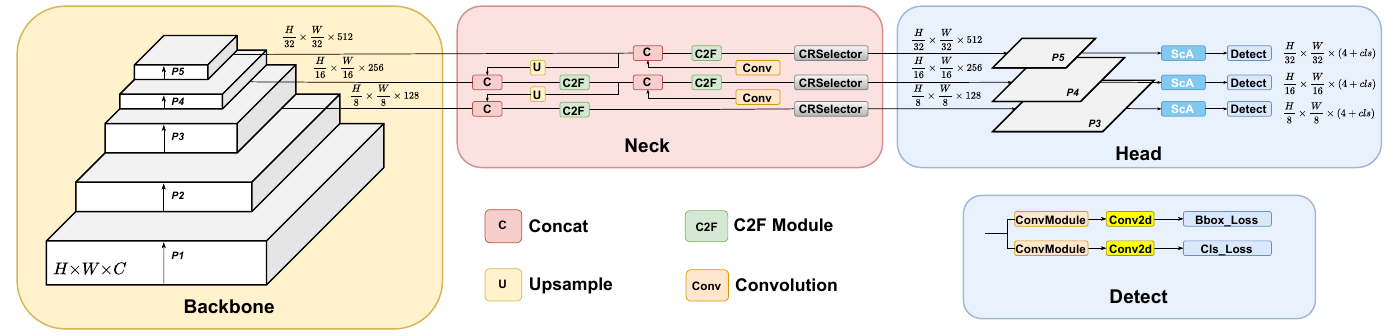}
	\caption{Overview of Fracture-YOLO, where CRSelector and ScA are improvements to the model based on the original.}
	\label{fig:fig2}
\end{figure*}
Traditionally, fracture diagnosis has relied on visual assessment by experienced radiologists. However, this manual assessment method relies heavily on the physician’s expertise and experience, limiting the speed of initial diagnosis. With advances in AI, data-driven fracture detection methods have shown great potential. Deep learning-based models can autonomously process and analyze large amounts of medical imaging data to provide a preliminary diagnosis and help doctors improve diagnostic efficiency. Therefore, the use of AI technology not only speeds up diagnosis and reduces the risk of misdiagnosis or missed diagnosis, but also enables rapid screening in emergencies, thus buying valuable treatment time. In recent years\cite{thian2019convolutional,krogue2020automatic,ahmed2024enhancing,zou2024detection,chien2024yolov9}, notable advancements have been achieved in fracture recognition using object detection, which allows real-time and accurate detection of fracture types.

However, in practical smart healthcare, object detection applications commonly face three challenges: poor imaging quality leading to indistinct features, fractures often appearing as subtle targets, and shape variability due to patient positioning. Poor imaging quality can result in the loss of edge features and texture details, causing the model to extract less representative features (see Fig. 1a). When both subtle and large targets are present in a scene, the model may struggle to capture features of all target sizes simultaneously (see Fig. 1b and 1c). Subtle targets within larger images are easily overlooked or appear blurred, limiting the model's capacity to effectively perform multi-scale feature extraction and recognition (see Fig. 1d). 

Additionally, patients are frequently positioned in varied postures during imaging, which may alter the appearance of the same target at different angles. This variation in posture can significantly change a fracture's shape, size, and position, making it challenging for conventional object detection models to recognize and process these shifts. In such cases, the model often captures only partial or blurred features, failing to adequately encompass the complete target information and ultimately diminishing detection performance.

In order to address the aforementioned challenges, a novel attention mechanism, referred to as Critical Region Selector Attention (CRSelector), is introduced \cite{wang2024camixersr}, with the aim of emphasizing critical regions for fracture detection. Additionally, an enhanced version of Scale-Aware Head (ScA) is incorporated to augment the model's capacity to detect objects at multiple scales \cite{dai2021dynamic}. These modules are embedded into the basic YOLOv8 architecture, resulting in a structurally optimized design tailored to fracture detection requirements.

Our contributions are summarized as follows:
\begin{itemize}
	\item A novel CRSelector module is incorporated into neck, which utilizes global texture information to identify important regions related to fractures, thereby enhancing the representation of fine fracture features.
	
	\item By incorporating CRSelector and integrating the ScA module into the detection head, a novel Fracture-YOLO model is proposed. The ScA module enhances the features of critical regions, improving the model’s adaptability to fractures of various sizes and imaging qualities. Both CRSelector and the ScA module have not been applied in fracture detection before; their integration significantly enhances detection performance, with no significant increase in computational time.
	
	\item Extensive experiments on public fracture detection datasets \cite{nagy2022pediatric} have shown that the Fracture-YOLO outperforms current state-of-the-art methods, demonstrating its potential in real-world applications.
\end{itemize}
\begin{figure*}[h]
	\centering
	\includegraphics[width=\textwidth]{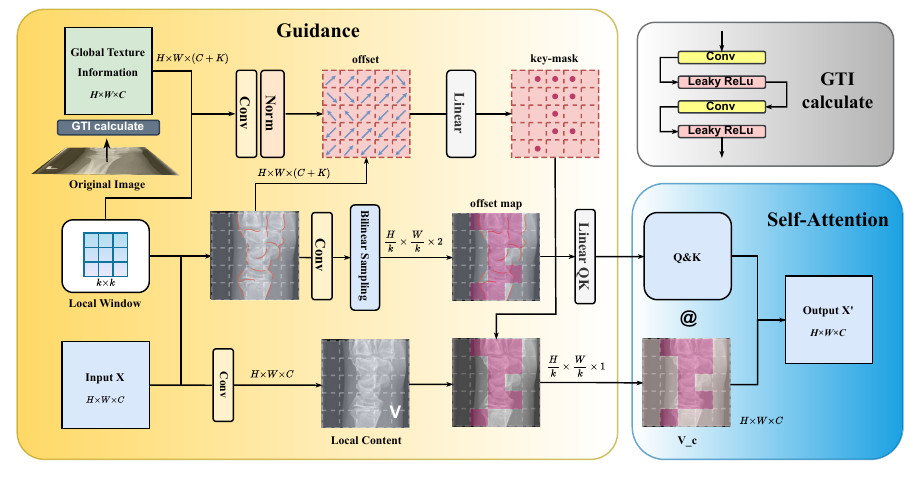}
	\caption{Overview of the CRSelector. CRSelector consists of two parts: Guidance and Self-Attention.}
	\label{fig:fig3}
\end{figure*}

\section{Related Works}
\subsection{Fracture Detection}
Fracture detection has been extensively studied, with deep learning models significantly improving diagnostic accuracy and efficiency. Thian \textit{et al.} \cite{thian2019convolutional} employed the Inception-ResNet Faster R-CNN architecture for the detection of wrist fractures, achieving sensitivities of 95.7\% and 96.7\% for frontal and lateral views. Krogue \textit{et al.} \cite{krogue2020automatic} used DenseNet for hip fracture classification with a binary accuracy of 93.7\%. Ahmed \textit{et al.} \cite{ahmed2024enhancing} evaluated YOLO-based models (YOLOv5–YOLOv8), reporting superior sensitivity (0.92) and mAP (0.95) over Faster R-CNN.
Zou \textit{et al.} \cite{zou2024detection} developed YOLOv7-ATT with attention mechanisms, achieving a mAP of 86.2\% on the FracAtlas dataset. Chien \textit{et al.} \cite{chien2024yolov9} introduced YOLOv9 for pediatric wrist fracture detection, improving mAP by 3.7\% compared to previous methods. While these studies have advanced fracture detection, challenges remain in handling complex patterns, low-quality imaging, and ensuring generalization.

\subsection{Attention Mechanisms}

Attention mechanisms enhance deep learning by focusing on relevant features. The Squeeze-and-Excitation (SE) block \cite{hu2018squeeze} recalibrates channel-wise features, achieving significant performance gains. Woo \textit{et al.} \cite{woo2018cbam} proposed CBAM, combining channel and spatial attention, offering consistent improvements across tasks. Wan \textit{et al.} \cite{wan2023mixed} introduced MLCA, balancing channel, spatial, local, and global texture information, improving mAP by 1.0\% over SE on Pascal VOC.
Huang \textit{et al.} \cite{huang2025generic} proposed DIA, which shares self-attention modules across layers and uses LSTM for calibration, reducing parameters while enhancing stability and performance. Ates \textit{et al.} \cite{ates2023dual} designed DCA to bridge the semantic gap in U-Net architectures by capturing multi-scale channel and spatial dependencies, improving segmentation performance with minimal parameter overhead.
Despite these advances, further exploration is needed to optimize attention mechanisms for diverse architectures and real-world applications.

\section{Method}
\subsection{Overall Architecture And Motivation}
As shown in the~Fig. \ref{fig:fig2}, the backbone of the detector adopts the CSPDarknet architecture\cite{terven2023comprehensive}, which effectively reduces the computational cost by introducing a branching mechanism, thereby improving the efficiency of image feature extraction. By integrating the SPPF layer, CSPDarknet has been further optimized, simplifying the model structure and enhancing overall inference speed.
The neck of the detector employs PAFPN \cite{zou2024detection} for feature fusion. It leverages bidirectional feature propagation (with bottom-up and top-down aggregation paths) to effectively utilize multi-scale information, enriching the feature maps and enhancing the model's adaptability to objects of various scales.
The YOLOv8 head directly generates groung-truth and class predictions from the feature map. The YOLOv8 head also retains the typical lightweight characteristics of the YOLO series, further enhancing model efficiency.
To address the challenges posed by poor imaging quality and the highly variable shapes of fractures, inspired by \cite{wang2024camixersr}, we have introduced a Critical Region Selector (CRSelector) at the neck of the model, along with a Scale-Aware (ScA) head. The CRSelector prioritizes critical regions by leveraging global texture features, while the Scale-Aware Head dynamically adjusts the attention weights based on different feature scales, enabling more accurate fracture detection under varying conditions.

\subsection{Critical-Region-Selector Attention}
In the context of fracture detection, medical image analysis focuses on identifying key features to locate the fracture. However, the presence of Unnecessary details and the neglect of subtle fracture regions in the feature hinder the accurate extraction of critical details, leading to inaccurate detection outcomes. This challenge is particularly pronounced for morphologically variable and multi-target fracture detection tasks, where incomplete or blurred features are often caused by variations in patient positioning and poor imaging quality.

To address these issues, a ritical-Region-Selector Attention (CRSelector) module has been designed (Fig. \ref{fig:fig3}), which aims to highlight the features of fracture regions by integrating both global and local semantic information, while suppressing redundant features. The CRSelector module is mainly composed of two elements: guidance and self-attention. The workflow of the module is as follows:

The module takes three inputs: the feature map 
$ \mathbf{X} $, the original image, and the local window. represents the feature map obtained from the backbone network and subsequently processed by the neck, the original image is the input, and the local window is predefined size. In subsequent computations, both the feature map $ \mathbf{X} $ and the original image will be divided into grid-like regions based on this size.

As shown in Fig. \ref{fig:fig3}, the guidance component produces two important outputs: the offset map and the key-mask, which are used to identify the critical regions of interest. Initially, the global texture information ($\mathit{GTI}$) is computed from the original image. This step involves extracting high-frequency features from the input which is required for detecting fracture details. Given the input, the $\mathit{GTI}$ is generated through two convolution layers. Based on the size of the local window, the $\mathit{GTI}$ and the $ \mathbf{X} $ are divided into multiple small regions, resulting in $\mathit{GTI}$ and $ \mathbf{X} $ containing location information.

Building on this, The input feature map $ \mathbf{X} $ is processed, and the local content $ \mathbf{V} $ is extracted using a $ 1 \times 1 $ convolution, which captures fine-grained details and serves to guide the subsequent calculations of the offset map and key-mask.

As shown in Fig. \ref{fig:fig3}, the offset map and key-mask are the core components of the guidance used to identify fracture regions \cite{wang2024camixersr}. The offset map achieves spatial deformation of the feature map by warping the window with more complex structures, ensuring that it aligns with the subtle structural changes in the fracture region. Specifically, the $\mathbf{offset}$ is first computed as follows:
\begin{equation}
    \mathbf{offset} = \mathit{tanh}\left(\text{Conv}_{1 \times 1}\left(\text{ReLU}\left(\mathit{Concat}(\mathbf{X},\mathit{GTI})\right)\right)\right) \cdot r,
\end{equation}
where $\mathit{Concat}$ refers to the concatenation of the two inputs ($\mathbf{X} $ and $\mathit{GTI}$), $\text{Conv}_{1\times1}$ represents the $1 \times 1$ convolution operation used to compress and transform the feature dimensions. The ReLU activation introduces non-linearity, and the $\mathit{tanh}$ function limits the offset values within the range of $[-1, 1]$. The scaling factor $r$ further adjusts the range of the predicted offsets. The resulting $\mathbf{offset} \in {\mathbb{R}}^\mathbf{2 \times H \times W}$ provides spatial displacement for each pixel, allowing the model to adaptively adjust the local window to better capture features and structural changes related to fractures.

After obtaining the offset, Bilinear interpolation is applied to the initial   $ \mathbf{X} $ to perform spatial deformation, shown as:
\begin{equation}
    {\mathbf{offset map}} = \sigma(\mathbf{X}, \mathbf{offset}),
\end{equation}
where $ \sigma(-) $ represents bilinear interpolation. The offset map emphasizes the specific features of the fracture, enabling the model to effectively capture cracks and irregular shapes.

The key-mask is obtained by computing the fusion of the local information $\mathbf{V} $ and the $\mathit{GTI}$. It represents the important regions after the local window division, expressed as a learnable probability distribution. The computation is as follows:
\begin{equation}
    \mathit{keymask} = \text{GumbelSoftmax}(f_{\text{reduce}}(\mathbf{V}\mathit{,GTI}) \mathbf{W}_{\text{mask}}),
\end{equation}
where $ f_{\text{reduce}} $ fuses $\mathbf{V} $ with $\mathit{GTI}$ and reduces the dimensionality, while $ \mathbf{W}_{\text{mask}} \in \mathbb{R}^\mathbf{M^2 \times 2} $ is a learnable weight matrix that maps the reduced features into a binary decision space. The Gumbel-Softmax function is used to binarize the computed mask to identify the regions of interest. This process involves applying the softmax function to the raw logits, normalizing the predicted probabilities within the range of $[0, 1]$. As a result, the probability of the important regions is higher, while the probability of irrelevant regions is lower, thereby highlighting the key areas.

After obtaining the offset map and key-mask, The $\mathit{key mask}$ is used to partition the deformed feature ($\mathbf{offset map}$) and the local information $\mathbf{V}$ into critical and normal regions:
\begin{equation}
    \left\{
    \begin{aligned}
	\mathbf{\tilde{K}} &= \mathbf{offsetmap} \otimes \mathit{keymask}, \\
	\mathbf{V}_{\text{c}} &= \mathbf{V} \otimes \mathit{keymask}, \\
	\mathbf{V}_{\text{n}} &= \mathbf{V} \otimes (1 - \mathit{keymask}),
    \end{aligned}
    \right.
\end{equation}
where $\otimes$ denotes element-wise multiplication (Hadamard product), the $\mathit{key mask}$ is the binary mask matrix computed earlier, and 1 - $\mathit{key mask}$ represents the complement of this matrix. $\mathbf{V}_{\text{c}}$ represents the critical region of the original feature map, while $\mathbf{V}_{\text{n}}$ represents the non-critical region.

By applying a linear transformation to the partitioned offset map K, The query ($ \tilde{\mathbf{Q}} $) and key ($ \tilde{\mathbf{K}} $) vectors are obtained:
\begin{equation}
    \tilde{\mathbf{Q}} = \tilde{\mathbf{K}} \mathbf{W}_q, \quad \tilde{\mathbf{K}} = \tilde{\mathbf{K}} \mathbf{W}_k.
\end{equation}

Self-attention is calculated based on the dot product of $ \tilde{\mathbf{Q}} $ and $ \tilde{\mathbf{K}} $  \cite{liu2021Swin}, and a weighted average is performed using the critical region $\mathbf{V}_{\text{c}}$. The process can be expressed as follows:
\begin{equation}
    \mathbf{X}' = \text{Conv}_{1 \times 1}\left(\text{softmax} \left( \tilde{\mathbf{Q}} \tilde{\mathbf{K}}^T \times d^{-\frac{1}{2}} \right) \mathbf{V}_{\text{c}}\right),
\end{equation}
where $ d $ is a scaling factor used to normalize the attention scores. This process assigns different levels of importance to the elements within the critical region, thereby highlighting their features.

Ultimately, the enhanced feature maps will be integrated with the original features, resulting in a refined representation of fracture regions that enables the model to more effectively detect key details in fractures with varying morphologies. The CRSelector module significantly improves detection accuracy and robustness, particularly in challenging medical imaging tasks, by effectively refining the feature maps through the integration process as shown in the equation:
\begin{equation}
    \mathbf{X} = \mathbf{X} + \mathbf{X}'.
\end{equation}

\subsection{Scale-Aware Head}
In prediction phase, features at different scales are crucial for capturing the diversity of fracture regions. Small cracks typically rely on low-level feature details, while large fracture areas or other related tissue require high-level semantic information. Therefore, selectively using multi-scale features is a key challenge in fracture target detection. To address this, a Scale-Aware Head (ScA) module is integrated to dynamically weight features at different scales, enabling the model to adaptively focus on scale-specific features.

\begin{figure}[h]
	\centering
	\includegraphics[width=\columnwidth]{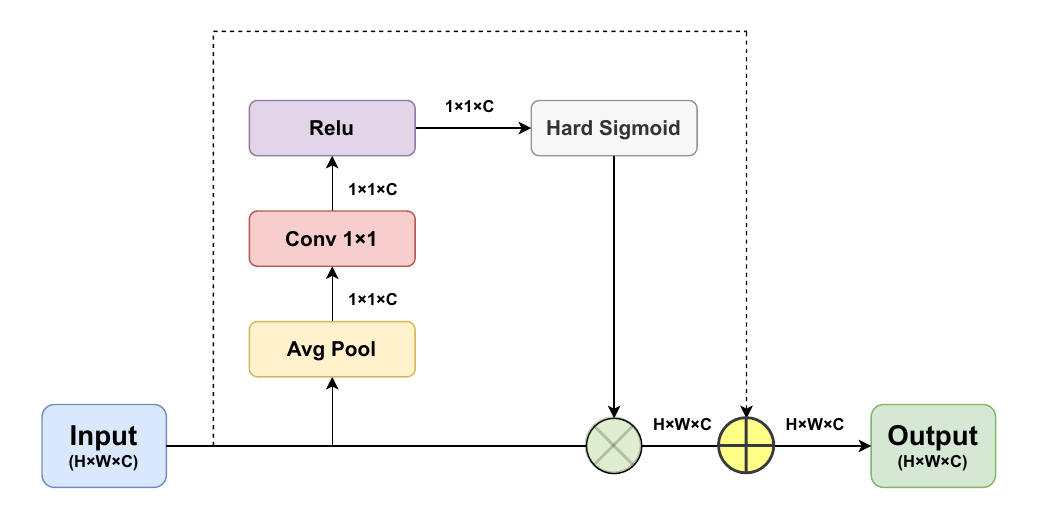}
	\caption{ The figure illustrates the architecture of the Scale-Aware (ScA) module, followed by a residual-style weighting operation to refine the feature representation.}
	\label{fig:fig4}
\end{figure}

The ScA module is shown in Fig. \ref{fig:fig4}. For an input feature tensor $\mathbf{F} \in \mathbb{R}^{H \times W \times C}$, it computes the scale-aware weights in a nested manner, which can be expressed as:
\begin{equation}
    \gamma_h = \phi\big(\text{ReLU}(\text{Conv}_{1 \times 1}(\mathbf{F}_{avg}(h)))\big),
\end{equation}
where $\mathbf{F}_{\text{avg}}(h) $ represents the global average feature for scale $ h $. This feature is processed by a $ 1 \times 1 $ convolution ($ \text{Conv}_{1 \times 1} $) followed by a ReLU activation to remove negative values and emphasize positive features. The result is passed through the hard sigmoid function $ \phi(x) = \max(0, \min(1, \frac{x+1}{2})) $, normalizing the attention value to the range $[0, 1]$. The final value of $ \gamma_h $ represents the importance weight for scale $ h $, allowing the model to adjust the feature contribution dynamically.

After obtaining the scale-aware attention weights $\gamma_h$, the input features are refined through a residual weighting operation. The refined feature representation is computed as:
\begin{equation}
    \mathbf{F}_{\text{weighted}} = \sum_{h=1}^H \left( \gamma_h \cdot \mathbf{F}_h + \mathbf{F}_h \right),
\end{equation}
where $\cdot$ denotes element-wise multiplication, and the addition introduces a residual connection, which aims to preserve the original input features and prevent the loss of important information in deep networks. Specifically, $\mathbf{F}_h$ represents the feature map of the $h$-th layer in the input feature tensor, and $\mathbf{F}_{weight}$ is the output feature. This process ensures the model  dynamically adjust the significance of features across various scales while retaining the original information. Through this process, the ScA module dynamically adjusts the contribution of features from different scales. For small cracks, it increases the weights of low-level features to capture fine details. For large and obvious fracture areas or to identify other issues, it emphasizes high-level features to focus on global information. By unifying multi-scale features, ScA greatly improves the model's capacity to identify fractures of different sizes and complexities.

Moreover, the ScA module incurs minimal computational overhead. This makes it easy to integrate into single-stage detectors (e.g., YOLO), enabling more efficient and accurate fracture detection.
\begin{table}[htbp]
	\centering
	\caption{GRAZPEDWRI-DX Dataset}
	\label{tab:1}
	\begin{tabular}{lcc} 
		\toprule
		\textbf{GRAZPEDWRI-DX} & \textbf{Train}  & \textbf{Test} \\
		\midrule
		Images          & 14,228 & 4,065   \\
		text            & 16,951 & 4,758  \\
		axis            & 14,228 & 4,065    \\
		fracture        & 12,675 & 3,635      \\
		periostealreaction & 2,391 & 724    \\
		pronatorsign    & 378   & 116      \\
		softtissue      & 326   & 101      \\
		metal           & 594   & 159      \\
		boneanomaly     & 203   & 48    \\
		bonelesion      & 31    & 10      \\
		foreignbody     & 4     & 2     \\
		\bottomrule
	\end{tabular}
\end{table}
\begin{table*}[htbp]
	\centering
	\caption{Comparison of the performance on GRAZPEDWRI-DX}
	\label{tab:2}
	\renewcommand{\arraystretch}{1.2} 
	\resizebox{\textwidth}{!}{ 
		\begin{tabular}{lcccccccc}
			\toprule
			\textbf{Methods} & \textbf{mAP(\%)} & \textbf{mAP$_{50}$(\%)} & \textbf{mAP$_S$(\%)} & \textbf{mAP$_M$(\%)} & \textbf{mAP$_L$(\%)} & \textbf{mAP$_{Fracture}$(\%)} & \textbf{Flops(G)} & \textbf{Parameters(M)}\\
			\midrule
			Vfnet~\cite{zhang2021varifocalnet} & 32.5 & 55.7 & 16.6 & 30.2 & 35.1 & 52.3 & 46.60 & 32.90 \\
			Retinanet~\cite{lin2017focal} & 25.1 & 46.5 & 12.0 & 21.3 & 27.7 & 50.4 & 50.31 & 36.52 \\
			Faster R-CNN~\cite{ren2015faster} & 28.6 & 53.7 & 12.6 & 20.7 & 31.7 & 49.7 & 60.49 & 32.89 \\
			Cascade R-CNN\cite{cai2018cascade} & 19.6 & 34.3 & 11.7 & 25.7 & 22.6 & 49.3 & 88.27 & 69.18 \\
                Mask R-CNN (Swin-T)~\cite{liu2021Swin} & 34.0 & 58.9  & 14.6 & 25.0 & 36.4 & 50.6 & 0.244 & 47.79 \\
			Mask R-CNN (Convnext)~\cite{liu2022convnet} & 37.2 & 62.5  & \textbf{19.0} & \textbf{35.5} & 37.9 & 51.6 & 0.238 &  47.72\\
			\midrule
			YOLOv5s  & 25.4 & 43.8 & 13.0 & 24.1 & 28.3 & 49.7 & 7.96 & 7.05\\
			YOLOv7tiny  & 31.7 & 55.1 & 14.0 & 29.3 & 33.7 & 51.8 & \textbf{6.59} & \textbf{6.04} \\
			YOLOXs~ & 36.9 & 63.9 & 17.8 & 31.9 & 38.3 & 53.8 & 13.33 & 8.94 \\
			YOLOv8s (Baseline) & 37.0 & 61.3 & 10.4 & 26.9 & 40.6 & 54.5 & 14.28 & 22.97 \\
			\midrule
			YOLOv8\cite{ju2023fracture} & 39.2 & 61.2 & - & - & - & - & 28.70 & 11.13 \\
			YOLOv8-ResCBAM\cite{ju2024yolov8} & 38.9 & 61.6 & - & - & - & - & 38.27 & 16.06 \\
			Ours & \textbf{40.0} & \textbf{65.3} & 15.1 & 34.6 & \textbf{41.7} & \textbf{55.5} & 17.99 & 34.80\\
			\bottomrule
		\end{tabular}
	}
\end{table*}

\begin{figure*}[htbp]
	\centering
	\includegraphics[width=15cm, height=9cm]{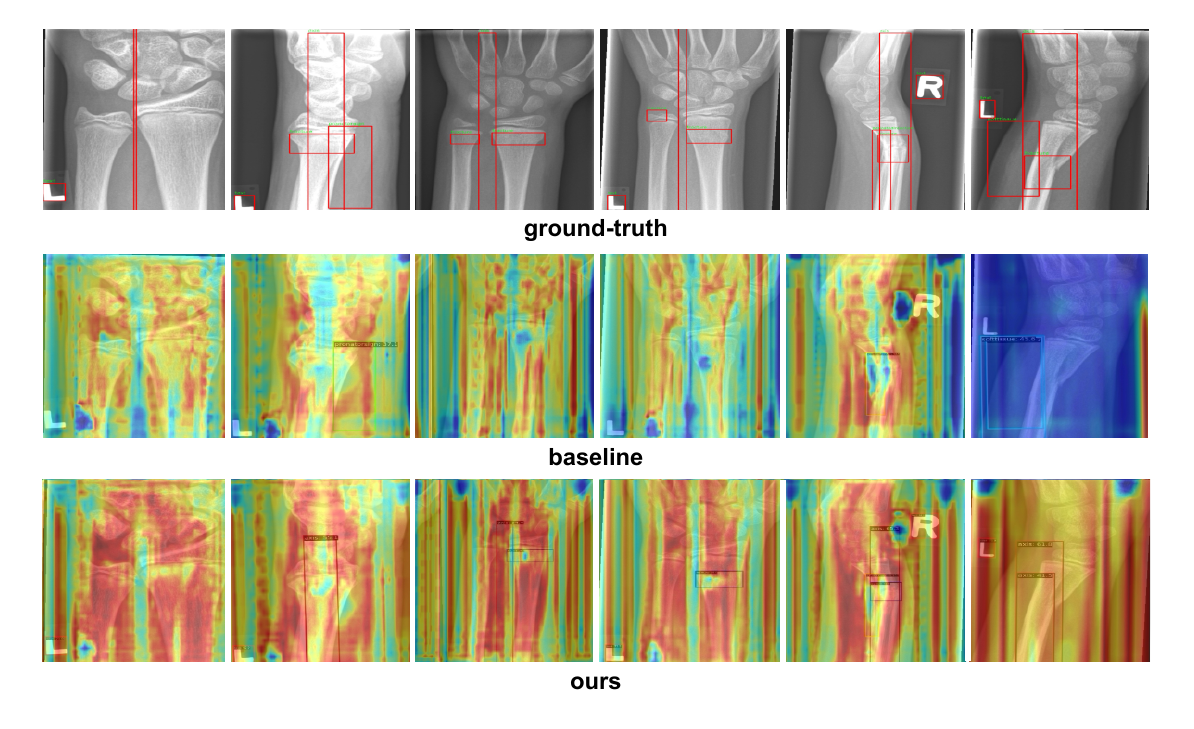}
	\caption{This figure compares the detection results between the baseline model (YOLOv8s) and our method in two challenging scenarios. The top row shows the original fracture images with ground-truth, where the red boxes represent detected fracture boxes. The middle row presents the heatmaps generated by the baseline model, and the bottom row shows the heatmaps generated by our method.}
	\label{fig:fig5}
\end{figure*}
\section{Experiment}
\subsection{Implementation Details and Dataset}\label{AA}  
All experiments were performed on a single NVIDIA RTX 4090 GPU. Our proposed Fracture-YOLO and other detectors are implemented using MMYOLO\cite{mmyolo2022} and MMDetection\cite{mmdetection}. Python version is 3.8, CUDA version is 11.8, Pytorch version is 2.0.0, torchvision version is 0.15.1. The proposed model is optimized with the SGD optimizer. During training, the learning rate is set to 0.01, the input image size is 640×640, and the batch size is 16. To reduce training time and computational costs, the model is initialized with pre-trained weights from the COCO dataset. All other models are evaluated with the same input image size and default parameters.
The dataset used, GRAZPEDWRI-DX \cite{nagy2022pediatric}, was provided by the University Medical School of Graz and contains 20,327 radiographic images of paediatric wrist trauma from 6091 patients between 2008 and 2018. As shown in Table \ref{tab:1}, the research leverages a comprehensive dataset characterized by its substantial breadth and depth, comprising 74,459 image labels and 67,771 meticulously annotated objects. This extensive collection encompasses a diverse spectrum of fracture classifications, ensuring robust representativeness.

\subsection{Evaluation Metrics}
In this experiment, the primary evaluation metric is mean Average Precision (mAP) \cite{lin2014microsoft}. Average Precision (AP) gauges the detection accuracy of a model by computing the mean precision across various recall levels, while mAP provides an overall assessment of the average precision across all categories. The mAP calculates the average accuracy over a range of Intersection over Union (IoU) thresholds from 0.5 to 0.95 \cite{lin2017focal}, and specifically provides \textit{mAP$_{50}$} results for an IoU threshold of 0.5. In addition, in order to more fully assess the performance of the model across different classes and object sizes, \textit{mAP$_S$}, \textit{mAP$_M$}, \textit{mAP$_L$}, and \textit{mAP$_{Fracture}$} are also reported for small, medium, large, and fracture objects. In addition to the accuracy metrics, the number of Flops (Floating Point Operations) is also reported, which is used as a measure of how many times the model has been propagated in a single forward propagation process. The number of model parameters is also reported to assess the computational complexity.
\begin{table}[htbp]
	\centering
	\caption{Effectiveness of CRSelector and ScA in YOLOv8s}
	\label{tab:3}
	\renewcommand{\arraystretch}{1.3} 
	\resizebox{0.5\textwidth}{!}{ 
		\begin{tabular}{l c ccc} 
			\toprule
			\textbf{Model} & \textbf{mAP(\%)} & \textbf{mAP$_{50}$(\%)} & \textbf{Flops(G)} & \textbf{Parameters(M)}\\
			\midrule
			YOLOv8s          & 37.0       & 61.3     & 14.28 & 22.97             \\
			+ CRSelector           & 39.1       & 63.4     & 17.38    & 33.87             \\
			+ ScA             & 38.2       & 63.1     & 14.89 &  23.9            \\
			Ours             & 40.0       & 65.3    & 17.99     & 34.80              \\
			\bottomrule
		\end{tabular}
	}
\end{table}
\subsection{Comparison Experiments}
\begin{table*}[htbp]
	\centering
	\caption{The performance of CRSelector and ScA in different models}
	\label{tab:4}
	\renewcommand{\arraystretch}{1.0} 
		\begin{tabular}{l c c c c} 
			\toprule
			\textbf{Methods} & \textbf{mAP(\%)} & \textbf{mAP$_{50}$(\%)} & \textbf{Flops(G)} & \textbf{Parameters(M)} \\
			\midrule
			Faster R-CNN     & 28.6             & 53.7                   & 60.49              & 32.89 \\
			+CRSelector            & 28.9             & 54.4                   & 63.50              & 43.79 \\
			+CRSelector+ScA         & 29.2    & 54.6         & 64.11              & 44.72\\
			\midrule
			YOLOv5s          & 25.4             & 43.8                   & 7.96               & 7.05\\
			+CRSelector            & 25.5             & 44.2                   & 11.06              & 17.95 \\
			+CRSelector+ScA         & 27.4    & 46.9          & 11.67              & 18.88 \\
			\midrule
			YOLOXs            & 36.9             & 63.9                   & 13.33              & 8.94 \\
			+CRSelector            & 37.2             & 63.6                   & 16.43              & 19.84 \\
			+CRSelector+ScA         & 37.2    & 64.3          & 17.04              & 20.77 \\
			\midrule
			YOLOv8s          & 37.0             & 61.3                   & 14.28              & 22.97 \\
			+CRSelector            & 39.1             & 63.4                   & 17.38              & 33.87 \\
			+CRSelector+ScA         & 40.0    & 65.3          & 17.99              & 34.80 \\
			\bottomrule
		\end{tabular}%
\end{table*}
Table \ref{tab:2} lists the results of our proposed method compared to the other detectors including Vfnet(ResNet-50)\cite{zhang2021varifocalnet}, Retinanet(Resnet-50)\cite{lin2017focal}, Faster R-CNN(Resnet-50)\cite{ren2015faster}, Cascade R-CNN(ResNet-50)\cite{cai2018cascade}, YOLOv5s, YOLOv7tiny, YOLOXs, YOLOv8\cite{ju2023fracture}, and YOLOv8-ResCBAM\cite{ju2024yolov8} on GRAZPEDWRI-DX. Our method demonstrates a significant lead, achieving impressive results of 65.2\% in \textit{mAP$_{50}$}, 40.0\% in \textit{mAP$_{50-95}$} and 55.5\% in \textit{mAP$_{Fracture}$} outperforming most existing approaches. Compared to YOLOv8s (baseline), our method achieves improvements of 4 in \textit{mAP$_{50}$}, 3 in \textit{mAP$_{50-95}$} and 1 in \textit{mAP$_{Fracture}$}, with a reduction of approximately 10.9M Parameters and 3.1G in FLOPs. Additionally, when compared to Cascade R-CNN (ResNet-50)\cite{cai2018cascade}, our method reduces the parameter count by approximately 34.38M. In addition, our method achieves a leading performance of 15.1\% in small object detection performance (\textit{mAP$_S$}), and 34.6\% and 41.7\% in medium object detection (\textit{mAP$_M$}) and large object detection (\textit{mAP$_L$}), respectively.
However, compared to Mask R-CNN (Convnext)~\cite{liu2022convnet}, our method slightly lags in \textit{mAP$_S$} 
and \textit{mAP$_M$} but outperforms it in all other metrics, while also requiring fewer parameters and computational resources.

\subsection{Ablation Experiments}
To further assess the effectiveness of the proposed method, each module was added separately to the baseline model, and their effects on mAP and Flops were evaluated. As shown in Table \ref{tab:3}, each module individually significantly improves the mAP of the baseline, while the best performance is achieved when combining the two modules. In addition, the CRSelector and ScA modules are integrated into four other detection models, and comparison experiments are conducted, the results are shown in Table \ref{tab:4}. The results show that, despite the acceptable increase in model complexity and number of parameters after the introduction of the modules, the mAP of most detection models shows an improvement, which further verifies the validity of our approach.
\begin{figure}[h]
	\centering
	\includegraphics[width=1\columnwidth]{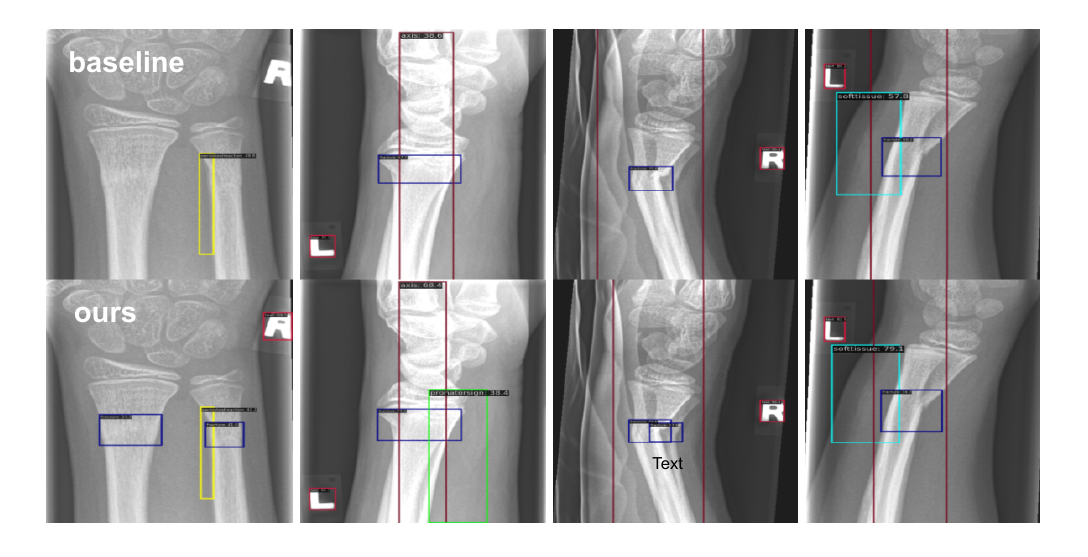}
	\caption{Comparison between Fracture-YOLO and the baseline model (YOLOv8s) for detection.}
	\label{fig:fig6}
\end{figure}
\subsection{Visualization}
To further assess the effectiveness of our method, A visual comparison of the detection results between the proposed method and Fracture-YOLO is presented. Feature maps from layers 0 and 1 are focused on, with Grad-CAM\cite{mmyolo2022} being leveraged to visualize these features. As shown in Fig. \ref{fig:fig5}, the heatmaps highlight the differences in feature attention between our model and the baseline. Our method significantly enhances the features of critical regions, particularly those that are small, difficult to detect, or occur in tissue covered areas.

As shown in Fig. \ref{fig:fig6}, for relatively simple fractures, although both models are able to detect fracture regions, the baseline model often fails to generate sufficiently accurate predict results due to inadequate feature extraction. By enhancing critical regions, our model significantly reduces these errors and generates independent and accurate predict results, thereby significantly improving detection precision.

\section{Conclusion}
In this paper, \text{Fracture-YOLO} is proposed, an enhanced fracture detection method that combines Critical-Region-Selector Attention (CRSelector) and Scale-Aware Heads (ScA) to effectively address the challenges of detecting small, subtle fractures and the morphological variations in medical imaging. Comprehensive experiments and visual comparisons show that our method significantly surpasses the baseline model across various fracture detection scenarios.
Compared to the baseline model, \text{Fracture-YOLO} improves detection accuracy and robustness, with mAP$_{50}$ and mAP$_{50-95}$ increasing by 4 and 3, respectively, achieving the current state-of-the-art (SOTA) performance. These results highlight the potential of \text{Fracture-YOLO} in handling complex fracture detection tasks and provide a reliable solution for clinical applications requiring precise fracture localization. Future research will aim to further refine feature extraction techniques to improve the model's performance across various clinical environments.
\bibliographystyle{IEEEtran}
\bibliography{IEEEexample}

\begin{thebibliography}{10}
\providecommand{\url}[1]{#1}
\csname url@samestyle\endcsname
\providecommand{\newblock}{\relax}
\providecommand{\bibinfo}[2]{#2}
\providecommand{\BIBentrySTDinterwordspacing}{\spaceskip=0pt\relax}
\providecommand{\BIBentryALTinterwordstretchfactor}{4}
\providecommand{\BIBentryALTinterwordspacing}{\spaceskip=\fontdimen2\font plus
\BIBentryALTinterwordstretchfactor\fontdimen3\font minus \fontdimen4\font\relax}
\providecommand{\BIBforeignlanguage}[2]{{%
\expandafter\ifx\csname l@#1\endcsname\relax
\typeout{** WARNING: IEEEtran.bst: No hyphenation pattern has been}%
\typeout{** loaded for the language `#1'. Using the pattern for}%
\typeout{** the default language instead.}%
\else
\language=\csname l@#1\endcsname
\fi
#2}}
\providecommand{\BIBdecl}{\relax}
\BIBdecl

\bibitem{shaik2024survey}
T.~Shaik, X.~Tao, L.~Li, H.~Xie, and J.~D. Vel{\'a}squez, ``A survey of multimodal information fusion for smart healthcare: Mapping the journey from data to wisdom,'' \emph{Information Fusion}, vol. 102, p. 102040, 2024.

\bibitem{mridha2023interpretable}
K.~Mridha, M.~M. Uddin, J.~Shin, S.~Khadka, and M.~F. Mridha, ``An interpretable skin cancer classification using optimized convolutional neural network for a smart healthcare system,'' \emph{IEEE Access}, vol.~11, pp. 41\,003--41\,018, 2023.

\bibitem{kothamali2023smart}
P.~R. Kothamali, N.~Srinivas, N.~Mandaloju, and V.~Karne, ``Smart healthcare: Enhancing remote patient monitoring with ai and iot,'' \emph{Revista de Inteligencia Artificial en Medicina}, vol.~14, no.~1, pp. 113--146, 2023.

\bibitem{thian2019convolutional}
Y.~L. Thian, Y.~Li, P.~Jagmohan, D.~Sia, V.~E.~Y. Chan, and R.~T. Tan, ``Convolutional neural networks for automated fracture detection and localization on wrist radiographs,'' \emph{Radiology: Artificial Intelligence}, vol.~1, no.~1, p. e180001, 2019.

\bibitem{krogue2020automatic}
J.~D. Krogue, K.~V. Cheng, K.~M. Hwang, P.~Toogood, E.~G. Meinberg, E.~J. Geiger, M.~Zaid, K.~C. McGill, R.~Patel, J.~H. Sohn \emph{et~al.}, ``Automatic hip fracture identification and functional subclassification with deep learning,'' \emph{Radiology: Artificial Intelligence}, vol.~2, no.~2, p. e190023, 2020.

\bibitem{ahmed2024enhancing}
A.~Ahmed, A.~S. Imran, A.~Manaf, Z.~Kastrati, and S.~M. Daudpota, ``Enhancing wrist abnormality detection with yolo: Analysis of state-of-the-art single-stage detection models,'' \emph{Biomedical Signal Processing and Control}, vol.~93, p. 106144, 2024.

\bibitem{zou2024detection}
J.~Zou and M.~R. Arshad, ``Detection of whole body bone fractures based on improved yolov7,'' \emph{Biomedical Signal Processing and Control}, vol.~91, p. 105995, 2024.

\bibitem{chien2024yolov9}
C.-T. Chien, R.-Y. Ju, K.-Y. Chou, and J.-S. Chiang, ``Yolov9 for fracture detection in pediatric wrist trauma x-ray images,'' \emph{Electronics Letters}, vol.~60, no.~11, p. e13248, 2024.

\bibitem{wang2024camixersr}
Y.~Wang, Y.~Liu, S.~Zhao, J.~Li, and L.~Zhang, ``Camixersr: Only details need more" attention",'' in \emph{Proceedings of the IEEE/CVF Conference on Computer Vision and Pattern Recognition}, 2024, pp. 25\,837--25\,846.

\bibitem{dai2021dynamic}
X.~Dai, Y.~Chen, B.~Xiao, D.~Chen, M.~Liu, L.~Yuan, and L.~Zhang, ``Dynamic head: Unifying object detection heads with attentions,'' in \emph{Proceedings of the IEEE/CVF conference on computer vision and pattern recognition}, 2021, pp. 7373--7382.

\bibitem{nagy2022pediatric}
E.~Nagy, M.~Janisch, F.~Hr{\v{z}}i{\'c}, E.~Sorantin, and S.~Tschauner, ``A pediatric wrist trauma x-ray dataset (grazpedwri-dx) for machine learning,'' \emph{Scientific data}, vol.~9, no.~1, p. 222, 2022.

\bibitem{hu2018squeeze}
J.~Hu, L.~Shen, and G.~Sun, ``Squeeze-and-excitation networks,'' in \emph{Proceedings of the IEEE conference on computer vision and pattern recognition}, 2018, pp. 7132--7141.

\bibitem{woo2018cbam}
S.~Woo, J.~Park, J.-Y. Lee, and I.~S. Kweon, ``Cbam: Convolutional block attention module,'' in \emph{Proceedings of the European conference on computer vision (ECCV)}, 2018, pp. 3--19.

\bibitem{wan2023mixed}
D.~Wan, R.~Lu, S.~Shen, T.~Xu, X.~Lang, and Z.~Ren, ``Mixed local channel attention for object detection,'' \emph{Engineering Applications of Artificial Intelligence}, vol. 123, p. 106442, 2023.

\bibitem{huang2025generic}
Z.~Huang, S.~Liang, and M.~Liang, ``A generic shared attention mechanism for various backbone neural networks,'' \emph{Neurocomputing}, vol. 611, p. 128697, 2025.

\bibitem{ates2023dual}
G.~C. Ates, P.~Mohan, and E.~Celik, ``Dual cross-attention for medical image segmentation,'' \emph{Engineering Applications of Artificial Intelligence}, vol. 126, p. 107139, 2023.

\bibitem{terven2023comprehensive}
J.~Terven, D.-M. C{\'o}rdova-Esparza, and J.-A. Romero-Gonz{\'a}lez, ``A comprehensive review of yolo architectures in computer vision: From yolov1 to yolov8 and yolo-nas,'' \emph{Machine learning and knowledge extraction}, vol.~5, no.~4, pp. 1680--1716, 2023.

\bibitem{liu2021Swin}
Z.~Liu, Y.~Lin, Y.~Cao, H.~Hu, Y.~Wei, Z.~Zhang, S.~Lin, and B.~Guo, ``Swin transformer: Hierarchical vision transformer using shifted windows,'' \emph{arXiv preprint arXiv:2103.14030}, 2021.

\bibitem{zhang2021varifocalnet}
H.~Zhang, Y.~Wang, F.~Dayoub, and N.~Sunderhauf, ``Varifocalnet: An iou-aware dense object detector,'' in \emph{Proceedings of the IEEE/CVF conference on computer vision and pattern recognition}, 2021, pp. 8514--8523.

\bibitem{lin2017focal}
T.-Y. Lin, P.~Goyal, R.~Girshick, K.~He, and P.~Doll{\'a}r, ``Focal loss for dense object detection,'' in \emph{Proceedings of the IEEE international conference on computer vision}, 2017, pp. 2980--2988.

\bibitem{ren2015faster}
S.~Ren, K.~He, R.~Girshick, and J.~Sun, ``Faster r-cnn: Towards real-time object detection with region proposal networks,'' \emph{Advances in neural information processing systems}, vol.~28, 2015.

\bibitem{cai2018cascade}
Z.~Cai and N.~Vasconcelos, ``Cascade r-cnn: Delving into high quality object detection,'' in \emph{Proceedings of the IEEE conference on computer vision and pattern recognition}, 2018, pp. 6154--6162.

\bibitem{liu2022convnet}
Z.~Liu, H.~Mao, C.-Y. Wu, C.~Feichtenhofer, T.~Darrell, and S.~Xie, ``A convnet for the 2020s,'' \emph{Proceedings of the IEEE/CVF Conference on Computer Vision and Pattern Recognition (CVPR)}, 2022.

\bibitem{ju2023fracture}
R.-Y. Ju and W.~Cai, ``Fracture detection in pediatric wrist trauma x-ray images using yolov8 algorithm,'' \emph{Scientific Reports}, vol.~13, no.~1, p. 20077, 2023.

\bibitem{ju2024yolov8}
R.-Y. Ju, C.-T. Chien, and J.-S. Chiang, ``Yolov8-rescbam: Yolov8 based on an effective attention module for pediatric wrist fracture detection,'' \emph{arXiv preprint arXiv:2409.18826}, 2024.

\bibitem{mmyolo2022}
M.~Contributors, ``Mmyolo: Openmmlab yolo series toolbox and benchmark,'' 2022, gitHub Repository.

\bibitem{mmdetection}
K.~Chen, J.~Wang, J.~Pang, Y.~Cao, Y.~Xiong, X.~Li \emph{et~al.}, ``Mmdetection: Open mmlab detection toolbox and benchmark,'' \emph{arXiv preprint}, vol. arXiv:1906.07155, 2019.

\bibitem{lin2014microsoft}
T.-Y. Lin, M.~Maire, S.~Belongie, J.~Hays, P.~Perona, D.~Ramanan, P.~Doll{\'a}r, and C.~L. Zitnick, ``Microsoft coco: Common objects in context,'' in \emph{Computer vision--ECCV 2014: 13th European conference, zurich, Switzerland, September 6-12, 2014, proceedings, part v 13}.\hskip 1em plus 0.5em minus 0.4em\relax Springer, 2014, pp. 740--755.

\end{thebibliography}

\end{document}